\documentclass[conference]{IEEEtran}
\usepackage{graphicx}
\usepackage{booktabs}
\usepackage{multirow}
\usepackage{cite}
\usepackage[noend]{algpseudocode}
\usepackage{algorithm}
\usepackage{algorithmicx}
\usepackage{color}
\usepackage{subfigure}
\begin{document}
\title{Semi-automatic Data Annotation System for Multi-Target Multi-Camera Vehicle Tracking}


\author{
    \IEEEauthorblockN{Haohong Liao$^{1,2}$, Silin Zheng$^{1,2}$, Xuelin Shen$^{2}$, Mark Junjie Li$^{1}$, Xu Wang$^{1*}$}
    \IEEEauthorblockA{1 College of Computer Science and Software Engineering, Shenzhen University, Shenzhen, China}
    \IEEEauthorblockA{2 Guangdong Laboratory of Artificial Intelligence and Digital Economy, ShenZhen, China}
    \IEEEauthorblockA{liaohaohong2021@email.szu.edu.cn, zhengpny@gmail.com, shenxuelin@gml.ac.cn, \{jj.li, wangxu\}@szu.edu.cn}
}

\maketitle

\begin{abstract}
Multi-target multi-camera tracking (MTMCT) plays an important role in intelligent video analysis, surveillance video retrieval, and other application scenarios. 
Nowadays, the deep-learning-based MTMCT has been the mainstream and has achieved fascinating improvements regarding tracking accuracy and efficiency.
However, according to our investigation, the lacking of datasets focusing on real-world application scenarios limits the further improvements for current learning-based MTMCT models.
Specifically, the learning-based MTMCT models training by common datasets usually cannot achieve satisfactory results in real-world application scenarios.
%
Motivated by this, this paper presents a semi-automatic data annotation system to facilitate the real-world MTMCT dataset establishment. 
The proposed system first employs a deep-learning-based single-camera trajectory generation method to automatically extract trajectories from surveillance videos.
Subsequently, the system provides a recommendation list in the following manual cross-camera trajectory matching process.
The recommendation list is generated based on side information, including camera location, timestamp relation, and background scene.
In the experimental stage, extensive results further demonstrate the efficiency of the proposed system.

\end{abstract}


\IEEEpeerreviewmaketitle

\section{Introduction}

Multi-target multi-camera tracking (MTMCT) is an important technology in intelligent transportation, city brain, intelligent video analysis, \textit{etc.} 
It aims to model vehicles' trajectories from multi-camera surveillance videos, contributing to vehicle tracking and traffic flow analysis and prediction.
%
The core of MTMCT is to compare the target vehicle from a certain camera view to a gallery of candidates captured by nearby cameras. 
%
In the past decade, the rapid development of machine learning technology shed new light on MTMCT studies. And the learning-based MTMCT has been the mainstream nowadays.
%
The learning-based MTMCT methods\cite{MCMOT,Ali,baidu,3th,AiCity} share a similar pipeline: the object detection and vehicle re-identification (re-ID) technologies are employed to detect the target vehicle and extract the corresponding features, respectively.
Then, the extracted features and the bounding boxes would be fed into multi-object tracking (MOT) module for trajectory generation, sequence-level clustering and cross-camera trajectory feature representation.
%

\begin{figure*}
    \centering
    \includegraphics[scale=0.5]{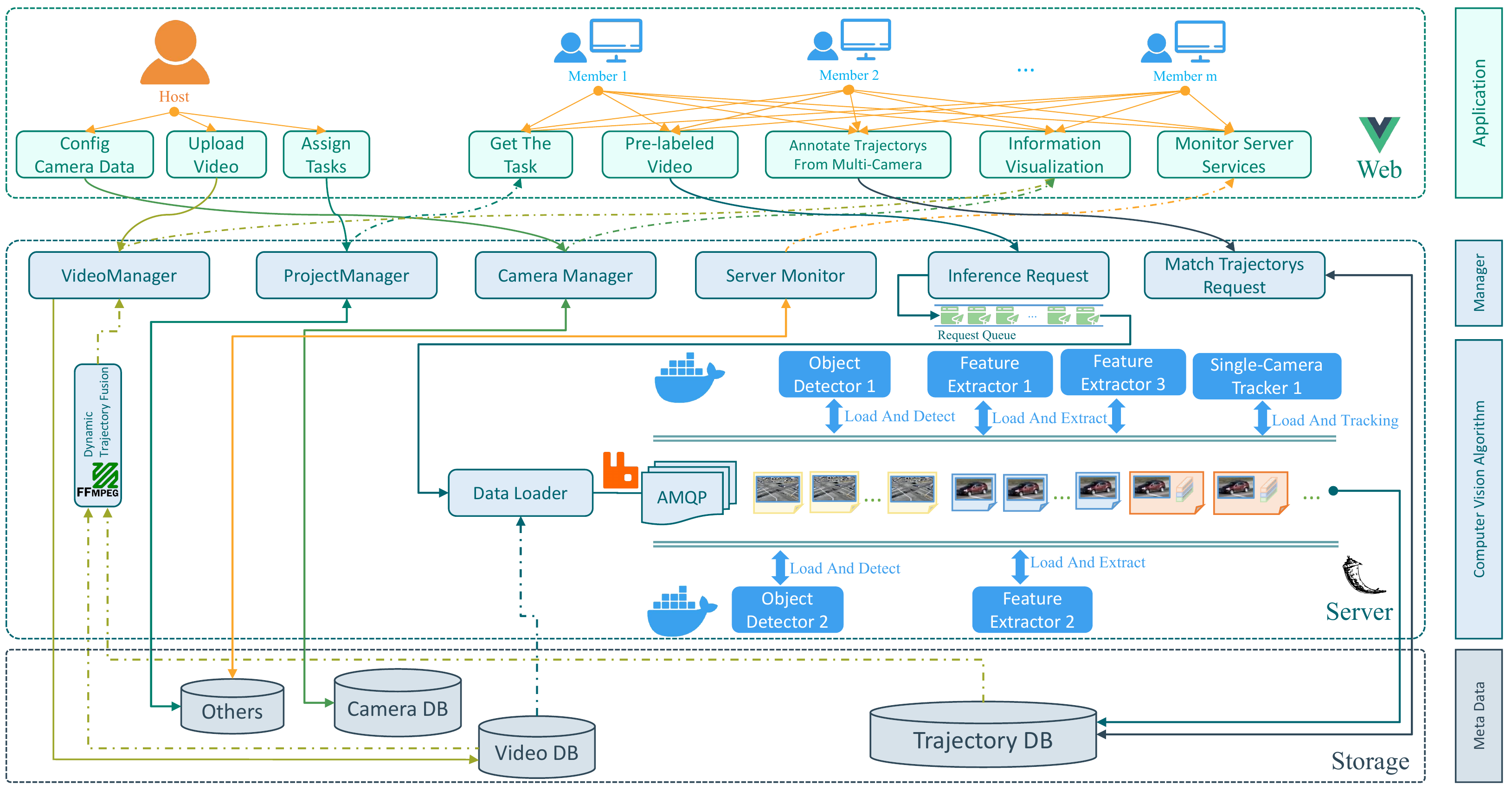} 
    \caption{The architecture of proposed system.
}
    \label{img_architecturel}
\end{figure*}

For the learning-based methods, the fundamental and indispensable work is establishing a steady and convincing dataset containing a wide variety of content and a sufficient number of trajectory annotations.
Unlike the datasets of common computer vision tasks, the MTMCT datasets do not only require labeling the targets within the video but also matching targets across videos from different cameras.
%
Conventional MTMCT datasets, such as CAVIAR4ReID\cite{CAVIAR4ReID}, WARD\cite{WARD} are constructed via manually labeling, which limits the dataset scale. 
Learning-based MTMCT models trained by these datasets usually cannot achieve satisfactory results. 
Recently, some works have leveraged  cutting-edge deep-learning technologies (such as ACF\cite{ACF} and Faster R-CNN\cite{fasterRCNN}) to realize automatic annotation.
Compared with the manually labeled datasets, which are usually at the thousand-level scales, the semi-automatic annotation method allows for establishing million-level scale datasets, such as MARS\cite{MARS}, MSMT17\cite{MSMT17}, \textit{etc.}

Although favorable achievements have been made in recent years, recent works \cite{reid_survey}\cite{semi-reid} have revealed the limitations caused by existing datasets.
%
%
Specifically, the learning-based MTMCT models trained by common datasets cannot achieve satisfactory results in real-world application scenarios due to the domain gap.
%
Under this circumstance, a feasible way is especially establishing a dataset according to the actual application scenario.
To this end, we propose a semi-automatic MTMCT data annotation system. 
To the best of our knowledge, it is the first semi-automatic annotation system focusing on the MTMCT dataset.
The architecture of proposed system is shown in Fig.~\ref{img_architecturel}.
%
%
%
The main contributions can be summarized as follows:
\begin{itemize}
    \item A deep-learning-based single-camera trajectory generation method is proposed in our system, where cutting-edge object detection and MOT models are employed to generate the vehicle trajectories automatically. After the automatic trajectory generation, the users will be asked to match the cross-camera vehicle trajectories. At this stage, the proposed system provides recommendation lists by making full use of side information (including camera location, timestamp, and background scene), significantly improving the manual labeling accuracy and efficiency.
    \item Regarding the system optimization, we propose a novel visualization strategy that dynamically generates the labeled videos via stored trajectory features.
    It saves plenty of storage requirements compared with the conventional method of directly saving the annotated video.
    Moreover, we employ the Linux Container technology and advanced message queuing protocol (AMQP) to improve the system's scalability.
    \item In the experimental stage, we validate the proposed system's efficiency based on the CityFlow\cite{cityflow} dataset. 
   Experimental results have demonstrated the accuracy of proposed deep-learning-based single-camera trajectory generation method, the efficiency of employed optimization strategy.
\end{itemize}

\section{Related Work}
\subsection{MTMCT}

\begin{figure}
    \centering
    \includegraphics[scale=0.35]{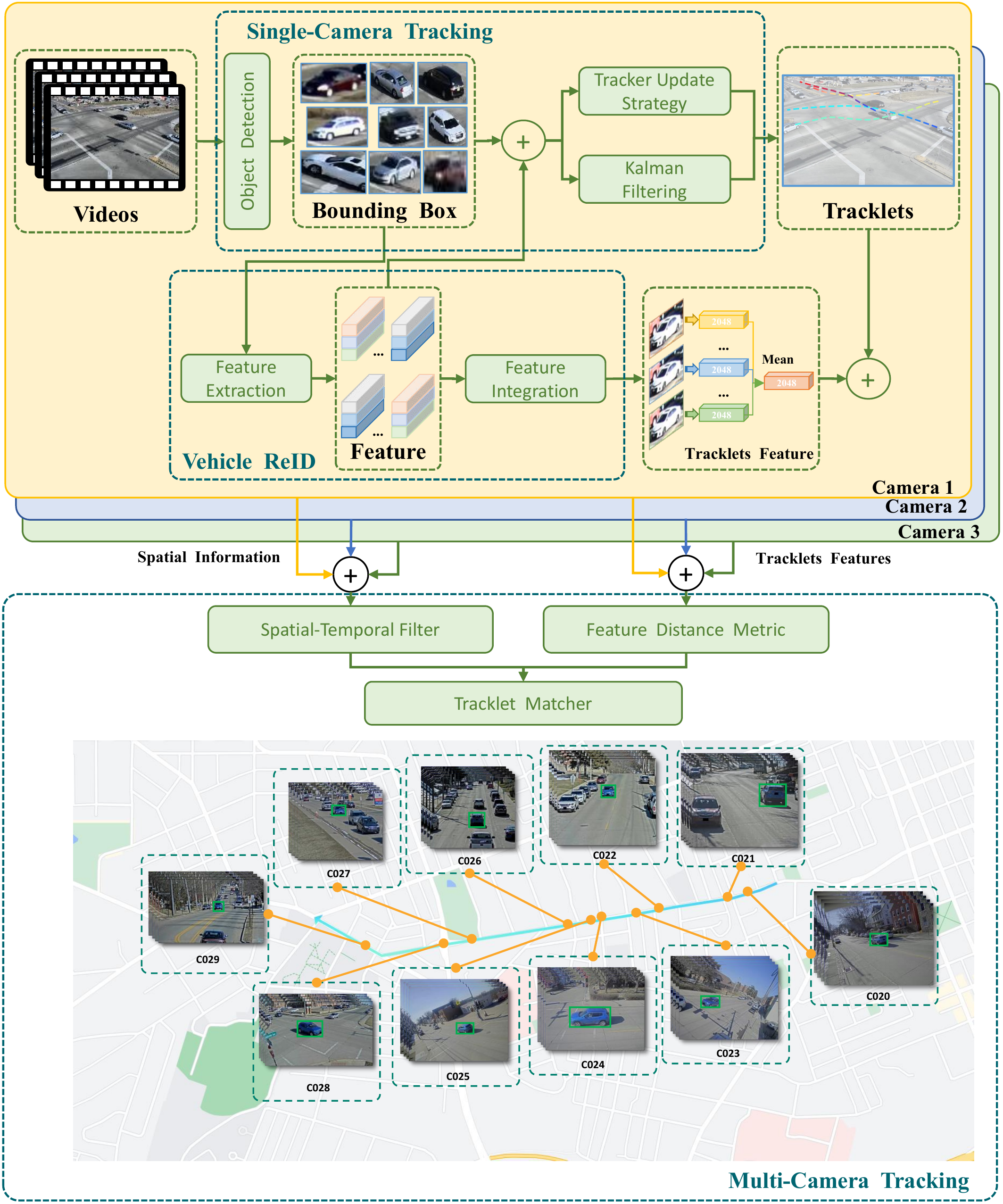}
    \caption{Illustration of Multi-Target Multi-Camera Tracking pipeline.}
    \label{MTMCT}
\end{figure}

The paradigm of MTMCT is illustrated in Fig.~\ref{MTMCT}. There are several sub-tasks wihtin the MTMCT pipeline:
(1) MOT, aims at extracting the trajectory feature vectors of the target vehicle from single-camera surveillance video.
(2) Vehicle re-ID, retrieves the target vehicle in a massive gallery set.
(3) Trajectory clustering, employs the spatial-temporal relation of related cameras to cluster target vehicle's cross-camera trajectory features.
In recent years, deep-learning-based MTMCT models have been rapidly developed. 
However, there are still many challenges.
In actual application scenarios, the occlusion, complex lighting conditions, and low video resolutions cause great trouble for MTMCT models.
Focusing on this, Liu \textit{et al.}~\cite{Ali} employs TFS filtering and DBTM constraint matching successfully increase the recall rate.
Ye \textit{et al.}\cite{baidu}
employ the TPM algorithm to 
overcome the influence caused by occlusion.

\begin{table*}
    \centering
    \caption{Summary of ReID dataset.}
    \begin{tabular}{cccccccccc}
        \toprule  
        Type    &Dataset    &Release Time   &Cams   &Imgs   &IDs    &Video  &Multiview  &Geom.  &Label Method   \\
        \midrule  
        \multirow{11}{*}{PersonReID}  &VIReR\cite{VIReR}  &2007   &2  &1,264  &632    &×  &×  &×  &Hand\\
        &GRID\cite{GRID}  &2009   &8  &1,275  &1,025    &×  &×  &×  &Hand\\
        &3DPes\cite{3DPes}  &2011   &8  &192  &1,011    &×  &$\surd$  &×  &Hand\\
        &PRID2011\cite{PRID2011}  &2011   &2  &1,134  &200    &$\surd$  &$\surd$  &×  &Hand\\
        &CUHK01\cite{CUHK01}  &2012   &2  &3,884  &971    &×  &$\surd$  &×  &Hand\\
        &CUHK02\cite{CUHK02}  &2013   &10  &7,264  &1,816    &×  &$\surd$  &×  &Hand\\
        &CUHK03\cite{CUHK03}  &2014   &10  &13,164  &1,467    &×  &×  &×  &DPM\cite{DPM}/Hand\\
        &Market-1501\cite{Market-1501}  &2015   &6  &32,668  &1,501    &×  &$\surd$  &×  &DPM\cite{DPM}/Hand\\
        &MARS\cite{MARS}  &2016   &6  &20,715  &1,261    &×  &$\surd$  &×  &DPM\cite{DPM}+GMMCP\cite{MCMOT}\\
        &DukeMTMC-reID\cite{DukeMTMC-reID}  &2017   &8  &36,411  &1,404    &×  &$\surd$  &×  &Hand\\
        &MSMT17\cite{MSMT17}  &2018   &15  &126,441  &4,101    &×  &$\surd$  &×  &Faster R-CNN\cite{fasterRCNN}\\
        \midrule
        \multirow{5}{*}{VehicleReID}  &VIRi-776 \cite{veri776}  &2016   &20  &49,357  &776    &×  &$\surd$  &$\surd$  &Hand\\
        &VehicleID\cite{vehicleid}  &2016   &2  &221,763  &26,267    &×  &$\surd$  &×  &Hand\\
        &VRAI\cite{VRAI}  &2019   &2  &137,613  &13,022    &×  &$\surd$  &×  &Hand\\
        &CityFlow\cite{cityflow}  &2019   &40  &229,680  &666    &$\surd$  &$\surd$  &$\surd$  &Object Detection+MOT\\
        &VeRi-Wild\cite{VeRi_Wild}  &2019   &174  &416,314  &40,671    &×  &$\surd$  &$\surd$  &YOLOv2\cite{yolo}\\
        \bottomrule 
    \end{tabular}
    \label{tab1}
\end{table*}

\subsection{Existing Dataset}
As exhibited in table~\ref{tab1}, we summarize the existing person and vehicle re-ID datasets from the aspects of dataset scale, target number, and camera diversity.
As shown, most of the existing datasets are based on manual labeling, which is extremely time-consuming~\cite{veri776,vehicleid,cityflow,VeRi_Wild,VRAI}.
As for VRAI~\cite{VRAI}, it spends around 1,000 hours to label the single vehicle bounding boxes, and 2,500 hours for cross-camera vehicle matching. 
%
%
Finally, 13,022 vehicle identifications are generated.
%
Recently, thanks to the rapid development of machine learning technology, deep-learning-based pre-labeling \cite{CUHK03,Market-1501,MARS,MSMT17,cityflow,VeRi_Wild} are proposed to save the resource requirement.
Take VeRi Wild~\cite{VeRi_Wild} as example, the vehicle bounding boxes are labeled via Yolov2\cite{yolo}, and 11 volunteers are invited to clean and organize 12 million original images for a month, finally collecting 40,671 vehicle identifications and 416,314 images.

Besides the datasets based on images,
in order to handle application scenarios requiring temporal domain information, datasets focusing on videos have been proposed, such as MARS\cite{MARS}, and CityFlow\cite{cityflow}.
%
%
However, the constructions of video-based datasets are extremely time-consuming since they require manual labeling frame by frame.
Under this circumstance, many deep-learning-based automatic labeling methods have been proposed\cite{activelearn1,activelearn2,virtuallearn1,virtuallearn2,semi-reid}.
As for \cite{activelearn1}\cite{activelearn2}, they employed reinforcement learning to achieve faster data iteration and train a re-ID network with manual loop supervision strategy.
In \cite{virtuallearn1}\cite{virtuallearn2}, the authors employed generative adversarial network (GAN) models to generate virtual image domain datasets focusing on digital city and digital twin. The dataset construction process barely requires manual labeling.
In \cite{semi-reid}, Zhao \textit{et.al.} proposed a semi-automatic labeling method for re-ID dataset. However, this method still requires plenty of manual data cleaning and sorting.

\section{The Proposed Semi-automatic Data Annotation System}
This section details the proposed semi-automatic MTMCT data annotation system. 
We first describe the deep-learning-based single-camera trajectory extraction.
Then, the provided manual matching assistance is detailed. 
At last, we present the efforts for system optimization regarding data visualization and adaptability.

\subsection{Deep-Learning-Based Single-Camera Trajectory Generation }
\label{sub:extract.trajectory}
Single-camera trajectory generation plays an important role in the proposed system, whose performance is directly related to the accuracy of the multi-camera matching.
Single-camera trajectory extraction mainly comprises two subtasks: MOT and trajectory feature generation.
%
%
%
As for a set of raw video sequences from different cameras, the proposed system will first sample key frames with a constant interval.
The YOLOv5\cite{yolo} algorithm is subsequently employed to detect all vehicles in each frame and return their bounding boxes. 
Then, features of detected vehicles will be extracted via a well-trained re-ID model.
Finally, the extracted features will be fed into the DeepSort\cite{DeepSort} algorithm for single-camera trajectory generation.
%
%
As for the other frames, their trajectories would be inferred by a linear interpolation method, which would save great storage and computation resources.
%
Specifically, denote the frame sample interval as $f$, and the bounding box of the frame $i$ as $b_i=(x_1^i,y_1^i,x_2^i,y_2^i)$. Then the bounding boxes of frame $j\in [I, i+f]$  can be inferred by the Eq.~(1), 

\begin{equation}
    b_j = b_i + \frac{j-i}{f} \times(b_{i+f}-b_i),
\end{equation}
Finally, following the method proposed in \cite{Ali}, the single-shot trajectory features from selected sequence is generated by,
\begin{equation}
    T = \frac{1}{N}\sum^N_{i=0}{f_i},
\end{equation}
where $N$ denotes the total frame number, and $f_i$ the feature vector of $i$-th. frame.
In the process of end-to-end trajectory extraction, we reduce the confidence threshold of the object detector to improve the recall rate of vehicle trajectories, but it will lead to a large number of false-positive samples. 
Therefore, we eliminate a large number of static targets and false detection samples according to the duration and the moving distances of the extracted trajectories.

\subsection{Manual Cross-Camera Trajectory Matching Assistance}
\label{sub:multi.structured}
\begin{figure}
    \centering
    \includegraphics[scale=0.4]{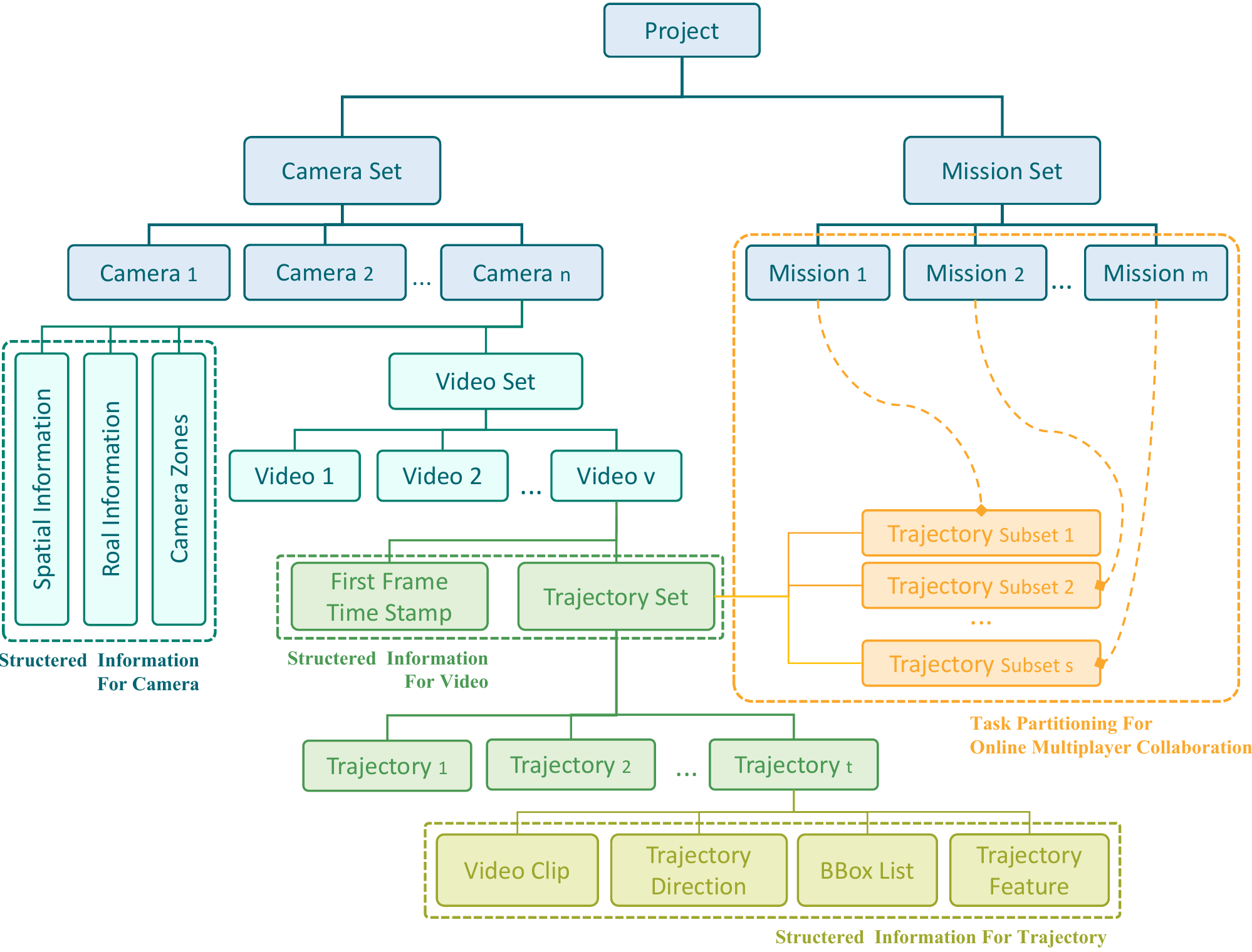}
    \caption{Employed side information for cross-camera  trajectory matching in the proposed system.}
    \label{structured.information}
\end{figure}

After the single-camera trajectory extraction, users will match trajectories from different cameras to construct the MTMCT dataset.
In this stage, the system provides a list of trajectories from nearby cameras for the target vehicle. 
In this process, the target occlusion, pose variance caused by shooting angle,  lighting condition variance, and poor video quality make it hard for the system to generate reliable recommendation lists.
Additionally, the similarities among the vehicles' shapes and colors usually cause great trouble for users' matching process.
%
%
Under these circumstances, the system makes full use of side information to establish reliable recommendation lists, and provide extra assistance for manual matching.
Fig.~\ref{structured.information} shows the employed side information.
This subsection details the generation of the side information.
%

\subsubsection{\textbf{Recommendation List Construction Based On Timestamp}}
\label{subsub:timestamp}

As for the target vehicle, the system would first provide a recommendation list for cross-camera matching.
The candidates are usually selected according to their appearance time-window in corresponding cameras' surveillance videos.
However, previous methods failed to consider the situation in actual application scenarios, where there are usually overlapping shoot areas among nearby surveillance cameras.
This usually causes incomplete recommendation lists, compromising the matching accuracy.
To this end, we employ Algorithm \ref{alg1} to get the maximum range of possible entry-time of the target under the downstream camera, 
and Algorithm \ref{alg2} to construct sorted gallery (CSG).
Take the two adjacent surveillance cameras $A$ and $B$ with overlapping shooting areas as an example, where A is in front of B.
%
The basic idea is that, the target's entry time-point of B is allowed being before or after the entry time-point of A, but must before the exit time-point of A.
%


\begin{figure}
    \centering
    \includegraphics[scale=0.25]{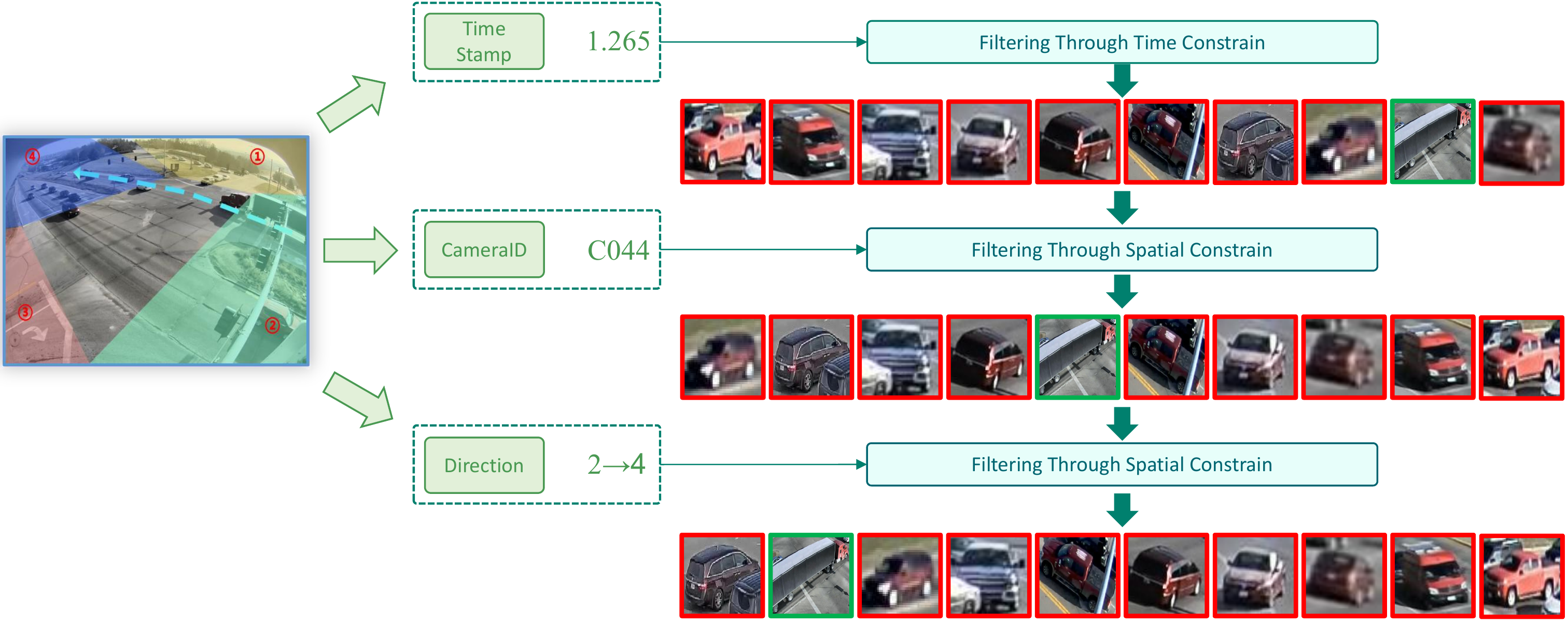}
    \caption{Illustration of recommendation list construction based on employed side information.}
    \label{information.filter}
\end{figure}

\subsubsection{\textbf{Recommendation List Cleaning based on Camera Location}}
\label{subsub:geom}
\begin{figure}
    \centering
    \subfigure[]{
    \includegraphics[scale=0.12]{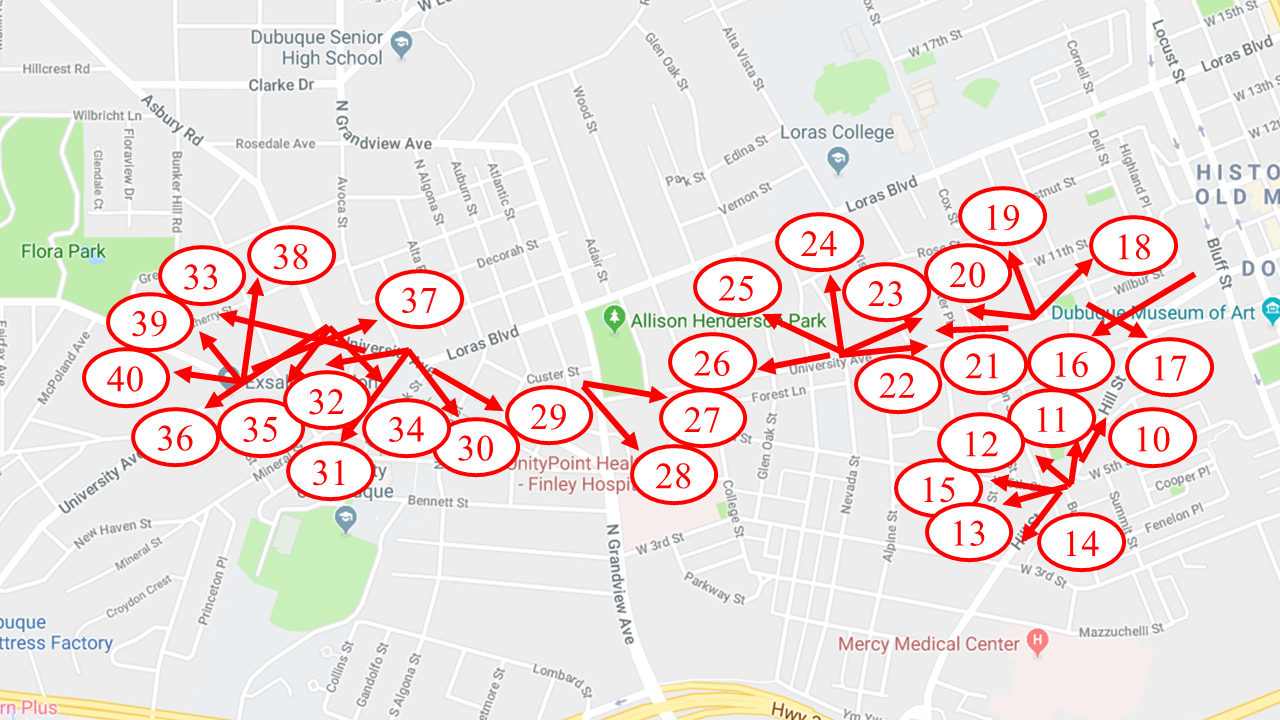}}
    \subfigure[]{
    \includegraphics[scale=0.021]{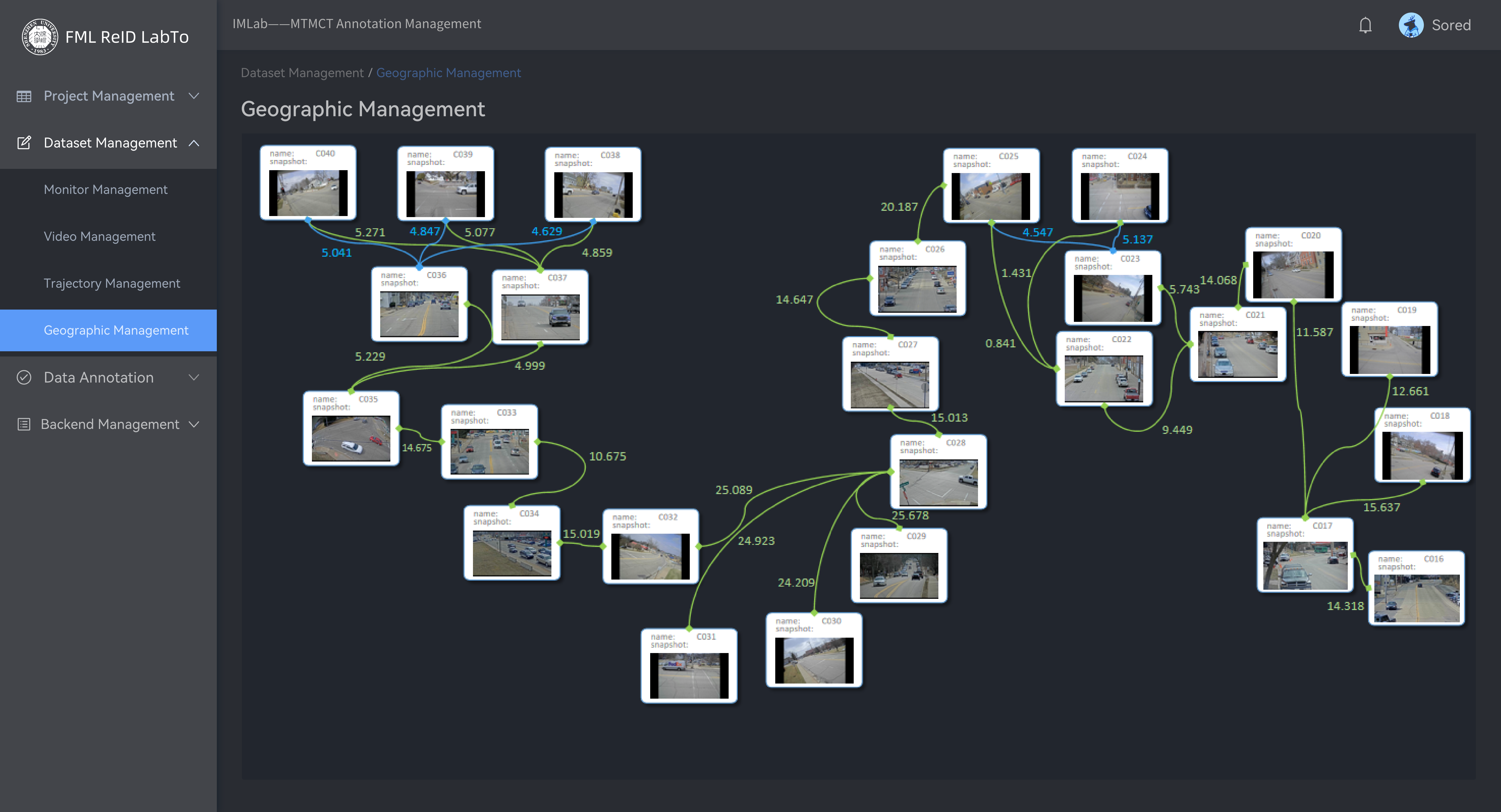}}
    \caption{Illustration of camera location information. (a) Camera location map in CityFlow; (b) Extracted camera location information.}
    \label{map}
\end{figure}

The cameras' location information will also be structured and employed to help construct the candidate list.
The basic idea is that, one vehicle's appearance time-point at corresponding cameras should be related to the cameras' locations~\cite{Ali}. 
Specifically, the target vehicle's single-camera-trajectory cannot be matched to trajectories from cameras that are too far away.
Besides, the basic traffic rules are also contributed to constructing the recommendation list.
Take the Fig.~\ref{information.filter} as an example, if the vehicle drives from the No.2 area to the No.4 area, trajectory candidates from cameras in No.1 and No.3 would be discarded.
Moreover, as shown in Fig.~\ref{map}, the structured camera location information is not only contributed to the candidate list construction but also be visualized and provided to users, which would be helpful in manual trajectory matching.
%
\begin{algorithm}[t]
    \caption{Construct Gallery Filter By Time Constrain}
    \label{alg1}
    \hspace*{0.02in}{\bf Require:} $C$(dict of camera),$O$(overlap time with query\\ camera),$st$(start time of trajectory),$et$(end time of trajectory)
    \hspace*{0.02in}{\bf Input:} $C_1$(list of gallery cameras), $Q$(trajectory in query set)\\
    \hspace*{0.02in}{\bf Output:} $G$(list of gallery after time constraint)
    \begin{algorithmic}[1]
        \For{$c \in C_1$}
            \If{area of c overlap with $C_2$}
                \State $st_{sch} \leftarrow st_{Q} - O$
            \Else
                \State $st_{sch} \leftarrow st_{Q}$
            \EndIf
            \State $g \leftarrow CSG(st_{sch}, G_c, Min, Max)$
            \State $G \leftarrow G \cup g$
        \EndFor\\
        \Return G
    \end{algorithmic}
\end{algorithm}

\begin{algorithm}[t]
    \caption{$CSG(st_{sch}, G_c, Min, Max)$}
    \label{alg2}
    \hspace*{0.02in}{\bf Require:} $j$(dict of single camera trajectory)\\
    \hspace*{0.02in}{\bf Input:} $st_{sch}$(st value of search space), $G_c$(list of trajectorys in gallery camera), Min(minimum search space), Max(maximize search space)\\
    \hspace*{0.02in}{\bf Output:} $g$(list of trajectorys in gallery)
    \begin{algorithmic}[1]
        \For{$j_g \in G_c$}
            \State $d \leftarrow st_j - st_{sch}$
            \If{$d \ge Min and d \le Max$}
                \State $add j_g to g$
            \EndIf
        \EndFor
        \State $g \leftarrow sorted(g)$\\
        \Return g
    \end{algorithmic}
\end{algorithm}

\subsubsection{\textbf{Side Information for Manual Matching Assistant}}
\begin{figure}
    \centering
    \includegraphics[scale=0.4]{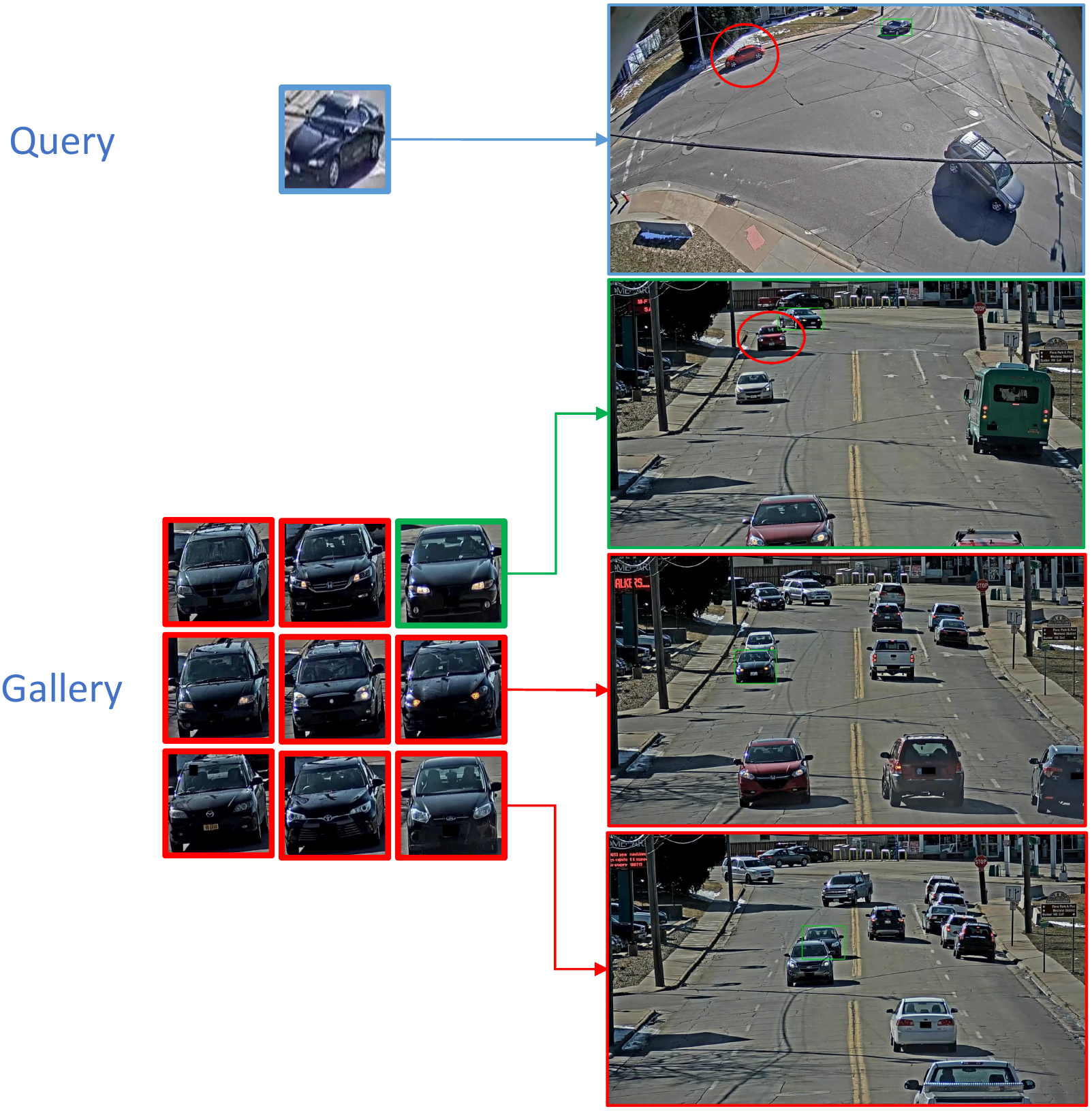}
    \caption{Employ the background scene from corresponding video clip as side information for cross-camera trajectory matching. }
    \label{bg_val}
\end{figure}

\begin{figure}
    \centering
    \includegraphics[scale=0.28]{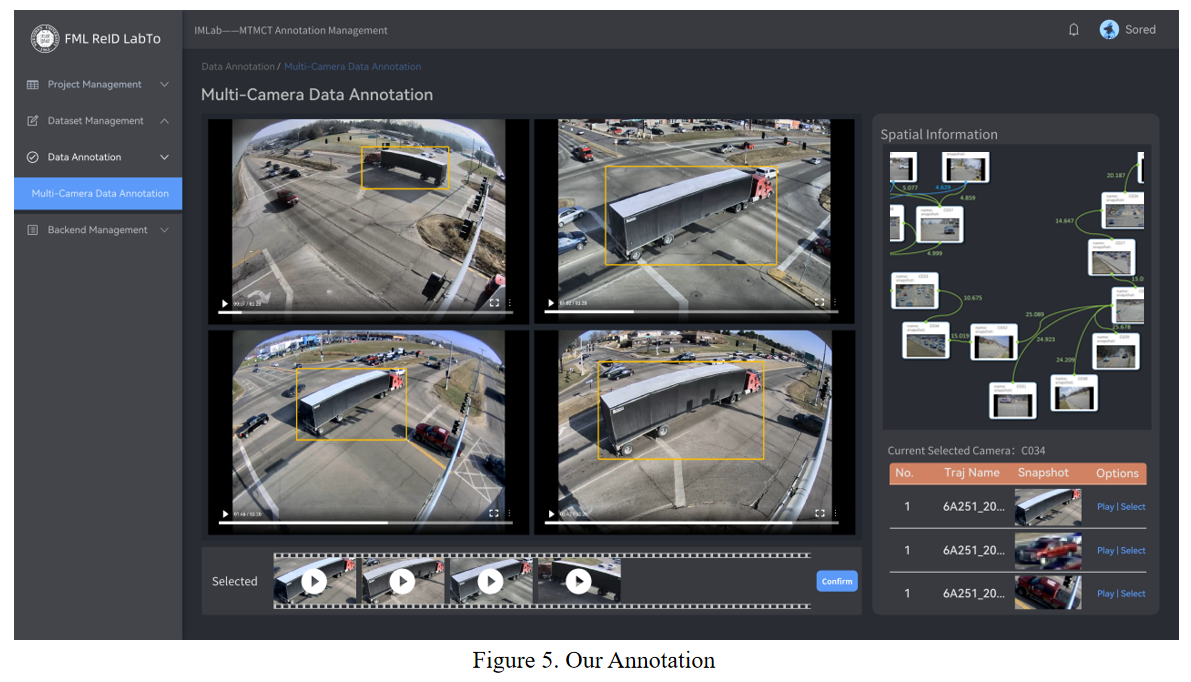}
    \caption{\textbf{The interface of our system.} The interface provide users visualized camera location information, recommendation list, video clips of recommended candidates.}
    \label{interface}
\end{figure}

The system also provides the original video clips of candidates to users, which greatly improves the matching accuracy in complex situations (such as indistinguishable vehicle shapes, occlusions, \textit{etc.}).
An example is shown in Fig.~\ref{bg_val}, users can easily distinguish similar vehicles according to the contained background scene with the provided video clips, avoiding matching errors.
A screenshot of the proposed system interface is shown in Fig.~\ref{interface}.
Specifically, the lower right part of the interface shows the recommendation list for trajectory matching,
%
the upper right part exhibits the camera location information of the whole area, 
and the left part displays the video clips of the candidate lists (four candidates' video clips can be displayed simultaneously).

\begin{figure*}[t]
    \centering
    \subfigure[]{
    \label{Fig.sub.1}
    \includegraphics[scale=0.55]{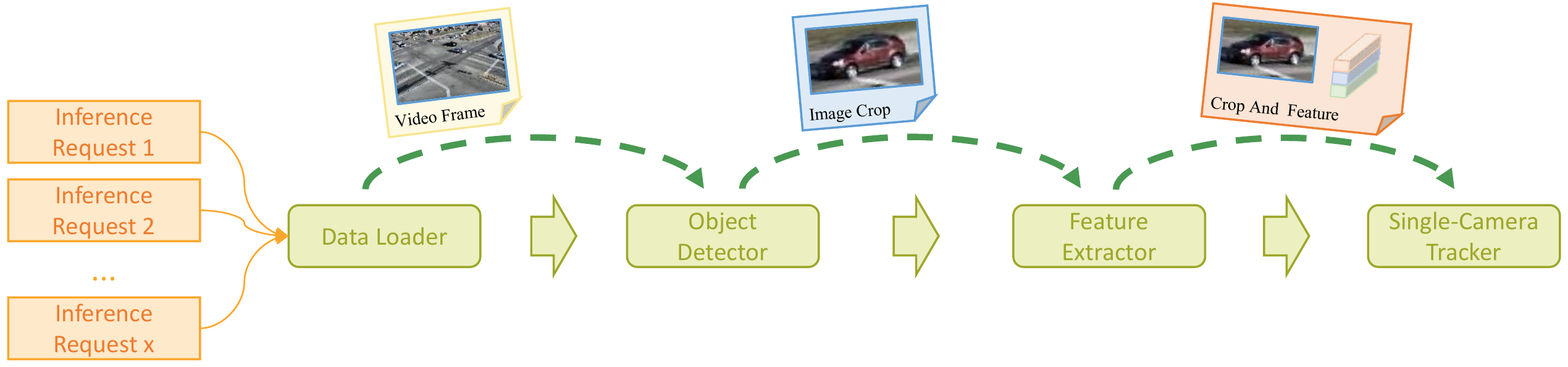}}
    \subfigure[]{
    \label{Fig.sub.2}
    \includegraphics[scale=0.55]{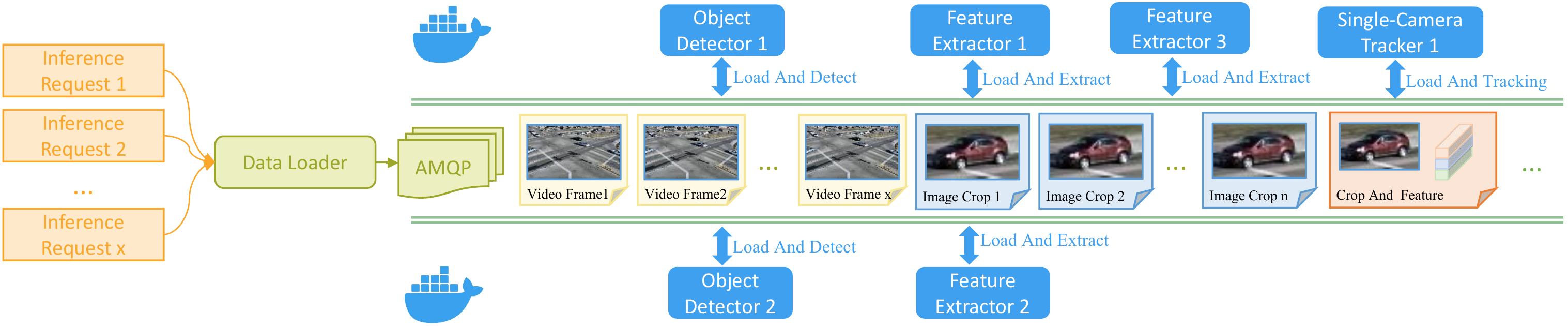}}
    \caption{Comparison between the traditional linear pipeline and the proposed system. (a) Traditional  linear  pipeline]; (b) Flexible deployment through Linux Container and advanced message queuing protocol (AMQP).} 
    \label{flexibledeploy}
\end{figure*}
After the cross-camera trajectory matching, the trajectory feature $T_i$ will be structured by,
\begin{equation}
    T_i = [(p_i, t_s, t_e, b_i), D_i, f_i],
\end{equation}
where $p_i$ is the original video data, $t_s, t_e$ are the starting time and the ending time of corresponding video clip, respectively; $b_i$ denotes the bounding box data of $T_i$; $D_i$ is the orientation of $T_i$ and $f_i$ is the corresponding feature vector of $T_i$.

\subsection{System Optimization Regarding Visualization and Adaptability.}
\label{sub:optimization.storage}
The existing re-ID datasets employ a similar visualization methodology that directly marks the trajectory data on the video sequence and stores the visualized trajectory data in video or GIF format.
This will cost plenty of storage resources.
Instead, the proposed system employs the FFmpeg toolbox to dynamically mark the trajectory data and visualize it via the web.
This allows the system to only store the video segments and trajectory features, saving plenty of storage and transfer resources.

We also devote plenty of effort to the system's adaptability and scalability.
Specifically, we employ the Linux Container to modularize different sub-tasks and AMQP for different modules' communication.
This strategy allows the proposed system to dynamically modularize the sub-tasks according to actual hardware environments, adapting to various application scenarios.
An intuitive illustration is shown in Fig.~\ref{flexibledeploy}.
%
%
In addition, the proposed system also supports multi-person online collaboration.
To achieve this, we employ a parallel processing strategy for the client-end and server-end.
The client-end employs the Vue framework to realize the visualization and interaction of the interface, and the server-end realizes dynamic data communication and semi-automatic algorithm inference through flask.
Furthermore, to improve the user experience of multi-user online collaboration and avoid blocking caused by simultaneous operating, the system implements asynchronous processing of requests via thread pools and request queue ranking mechanisms.

\section{Experimental Results}
In this section, we evaluate the proposed system from the aspect of trajectory generation accuracy, time complexity and annotation performance.
 %
Specifically, we will first introduce the evaluation criteria and the implementation details.
The experimental results and corresponding analysis will be subsequently presented.

\subsection{Implement Details and Evaluation Criteria}
\label{subEval}
As for the deep-learning-based trajectory generation evaluation, we roughly follow the method employed in ~\cite{fairmot}\cite{IDPer}.
%
Specifically, the true-positive identities (IDTP, the matching degree of extracted trajectory and ground-truth is more than 80\%), false-positive identities (IDFP), and false-negative identities (IDFN, the matching degree of extracted trajectory and ground-truth is less than 20\% or the system fails to generate the trajectory) would be first calculated, and subsequently contributed to the $Precision$ and $Recall$ computation,
%

\begin{equation}
    Precision = \frac{IDTP}{IDTP + IDFP},
\end{equation}
\begin{equation}
    Recall = \frac{IDTP}{IDTP + IDFN}.
\end{equation}
$Precision$ and $Recall$ count for the accuracy of deep-learning based trajectory generation.
Higher values mean more reliable trajectories generated by the proposed system.
%
The experimental environment is exhibited in table~\ref{tab2}. 
As for the object detection module, we employ the YOLOv5x\cite{yolo} model trained by COCO dataset.
The confidence threshold is adjusted to 0.1 to achieve high $Precision$ and $Recall$ values. 
In terms of single-camera MOT module, we adjust the time threshold to 1 second and the IoU threshold of the trajectory moving distance to 0.05 to filter the generated trajectories. 
%
The proposed system employs a parallel processing strategy of performing these modules on different GPUs to improve generation efficiency.
%
We test the deep-learning-based trajectory generation on the CityFlow\cite{cityflow} dataset, and the initial frame rate is set to 10 frame-per-second.
Things have to be mentioned that we discard videos barely containing trajectory data to avoid unnecessary errors in the experimental stage.

\begin{table}
    \centering
    \caption{Testing environment.}
    \begin{tabular}{cc}
    \toprule
    Name     &Performance  \\
    \midrule
    CPU     & Intel Xeon(R) Silver 4210R CPU@2.40GHz×40\\
    GPU     & Intel NVIDIA GeForce RTX-3090×2\\
    Memory     & 256GB\\
    Disk     & 4TB\\
    System     & Ubuntu18.04 LTS\\
    \bottomrule
    \end{tabular}
    \label{tab2}
\end{table}

\subsection{Experiment Result}
\label{subExp}

\subsubsection{\textbf{Evaluation of Deep-Learning-based Trajectory Generation}}
\label{subsub:alg}
%
The experimental results are shown in Table~\ref{tab3}, where 'S\#' denotes the scene number in CityFlow\cite{cityflow} and 'c\#' is the corresponding camera ID.
Quite encouraging results have been shown, that the proposed deep-learning-based trajectory generation's accuracy reaches 86.98\% and 86.81\% regarding $precision$ and $recall$ on average.
This provides strong evidence that the proposed system is capable of handling actual application scenarios. 
%
Moreover, the system also supports the users in changing the trajectory generation algorithm according to actual needs, which should be owed to the low-coupling modular design.

\begin{table}
    \centering
    \caption{Experimental Results of Deep-Learning-based Trajectory Generation.}
    \scalebox{0.9}{
    \begin{tabular}{cccccc}
    \toprule
        &Name     &Algorithm   &Ground-Truth &Precision (\%)  &Recall (\%)  \\
    \midrule
    \multirow{2}{*}{S01} &c003    &79 &71 &89.87  &100.00\\
    &c005    &80 &93 &90.00  &77.42\\
    \midrule
    \multirow{2}{*}{S02} &c006    &120 &123 &93.33  &91.06\\
    &c009    &125 &136 &92.00 &84.56\\
    \midrule
    \multirow{4}{*}{S04} &c030    &18 &23 &94.45  &73.91\\
    &c033    &19 &26 &100.00 &73.08\\
    &c034    &28 &24 &78.57 &91.67\\
    &c036    &22 &21 &77.27 &80.95\\
    \midrule
    \multirow{16}{*}{S05} &c010    &45 &28 &55.56  &89.29\\
    &c016    &84 &80 &84.52 &88.75\\
    &c017    &76 &84 &96.05 &86.90\\
    &c018    &55 &58 &94.55 &89.66\\
    &c021    &72 &67 &83.33 &89.55\\
    &c022    &84 &83 &97.62 &98.80\\
    &c023    &54 &83 &98.15 &63.84\\
    &c025    &69 &86 &97.10 &77.91\\
    &c026    &86 &89 &95.35 &92.13\\
    &c027    &77 &77 &85.71 &85.71\\
    &c028    &88 &69 &75.00 &95.65\\
    &c029    &119 &109 &70.59 &77.06\\
    &c033    &225 &204 &86.67 &95.59\\
    &c034    &192 &169 &83.85 &95.27\\
    &c035    &195 &190 &94.36 &96.84\\
    &c036    &137 &115 &73.72 &87.83\\
     \midrule
     &Average    &- &- &86.98 &86.81\\
    \bottomrule
    \end{tabular}
    }
    \label{tab3}
\end{table}

\subsubsection{\textbf{Evaluation Of the Proposed Trajectory Generation Method's Scalability}}
\label{subsub:labelTime}
\begin{figure}
    \centering
    \subfigure[]{\includegraphics[scale=0.17]{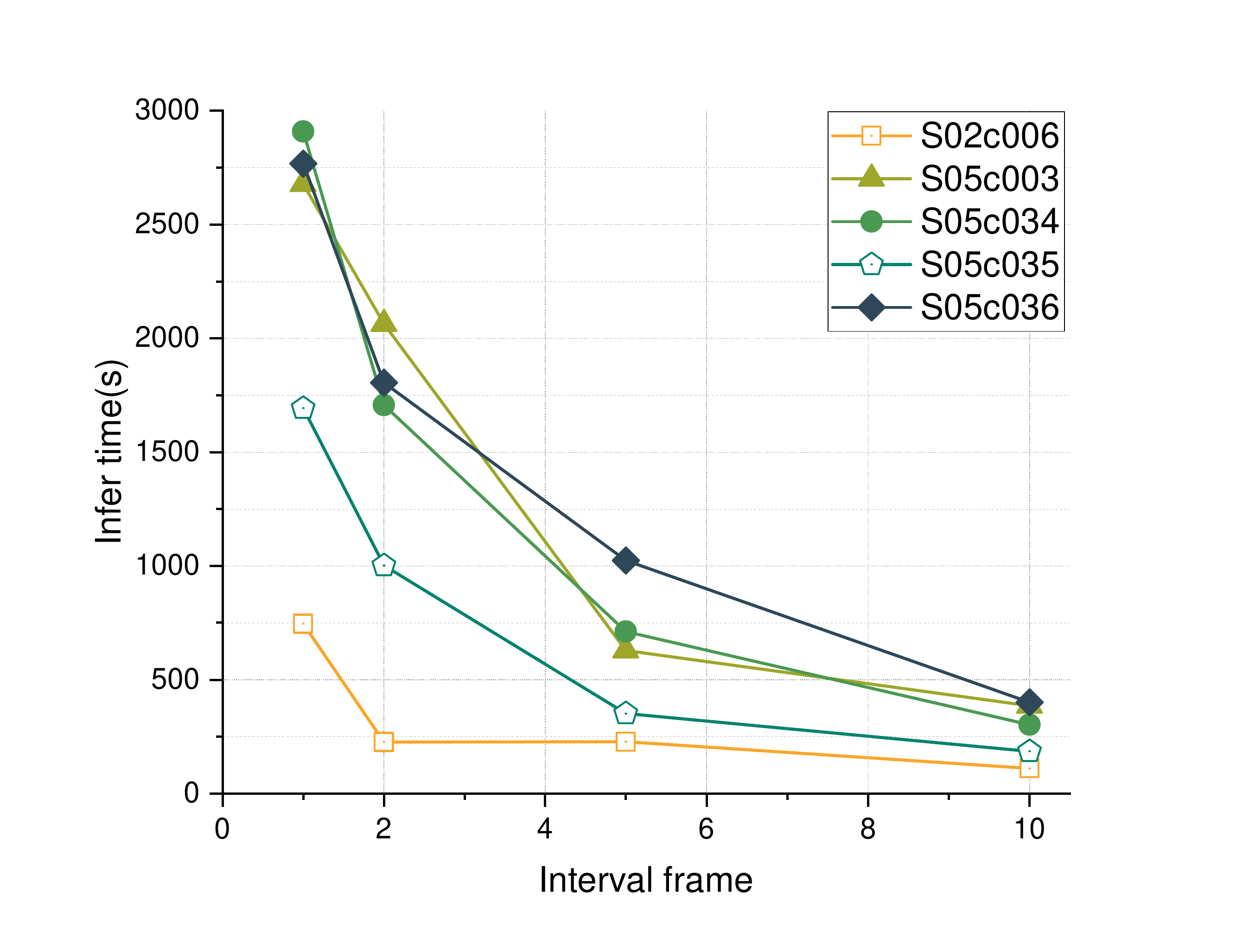}}
    \subfigure[]{\includegraphics[scale=0.17]{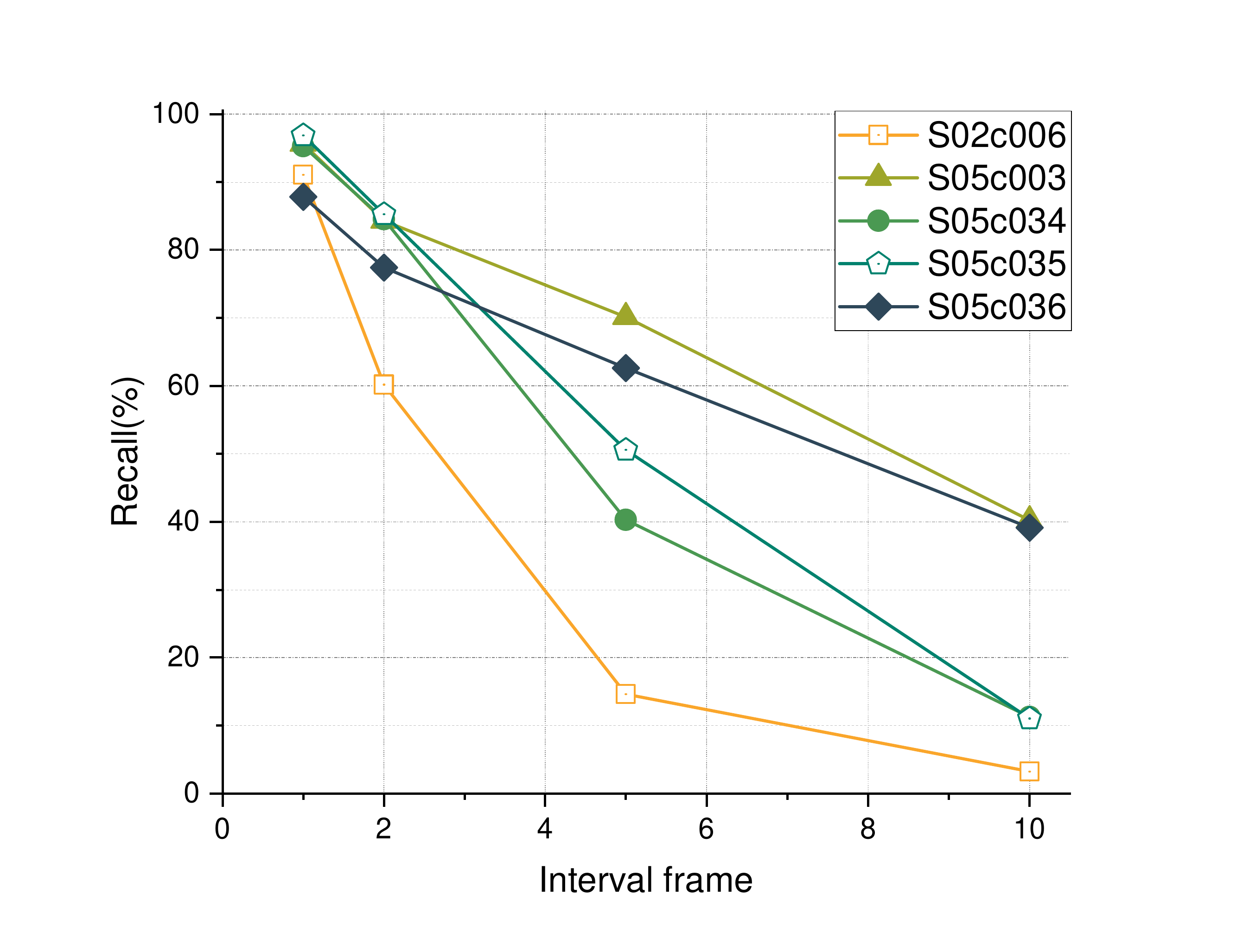}}
    \caption{\textbf{The influence of different parameters of frame extraction on the inference result.} (a) Inference Time (s); (b) Recall (\%). It can be seen from the line graphs (a) and (b) that as the parameter of frame extraction continues to increase, its running time and recall rate are constantly decreasing.}
    \label{tubiao_1}
\end{figure}

As mentioned in Sec.III, the proposed system provides a scalable setting.
Specifically, by adjusting the frame sample interval, users are allowed to have a trade-off between the running time and inferring accuracy.
This ensures the proposed system can be adapted to different application scenarios. 
Here, we conduct extensive experiments with different frame sample intervals to provide guidance for actual applications.
And the corresponding running time, $recall$ and extraction quality are reported in Table~\ref{tab4} and Fig.~\ref{tubiao_1}, where the 'Abnormal' refers to situations of failing to generate bounding boxes, extract target vehicles and extract trajectories. 
%
%
As is shown, when the sample interval is higher than 5, the system's performance drops significantly regarding
recall rate and the quality of the trajectory generation.
By adjusting the frame sample interval to 2, the system can achieve favorable performance with acceptable time complexity.  
%

\begin{table}
    \centering
    \caption{Experimental Results Of Scalable Settings.}
    \scalebox{0.7}{
    \begin{tabular}{cccccccc}
    \toprule
    Name     &Length (s)   &FPS &Interval   &Infer time (s) &Recall (\%) &Observe target  &Extract result\\
    \midrule
    \multirow{4}{*}{S02/c006} &\multirow{4}{*}{211} &\multirow{4}{*}{10} &1  &746   &91.06  &\multirow{4}{*}{\includegraphics[width=16mm, height=12mm]{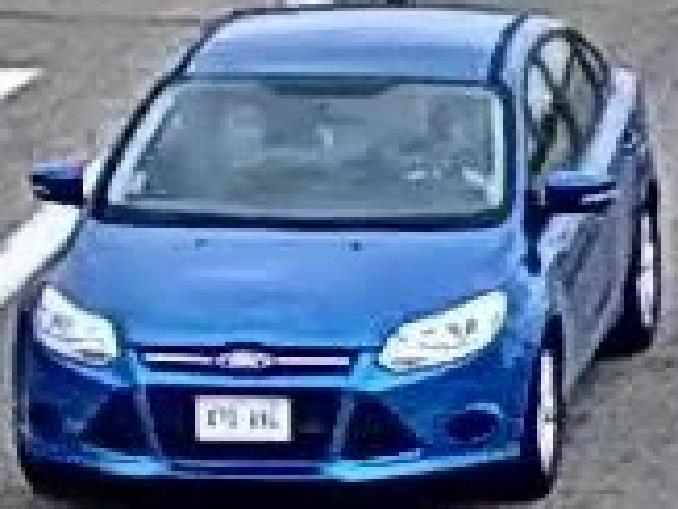}}   &Normal\\
        &   &   &\textbf{2}  &\textbf{226}   &\textbf{60.16}  &    &Normal\\
        &   &   &5  &227   &14.63  &    &Abnormal\\
        &   &   &10  &111   &3.25  &    &Abnormal\\
    \midrule
    \multirow{4}{*}{S05/c033} &\multirow{4}{*}{340} &\multirow{4}{*}{10} &1  &2,677   &95.59  &\multirow{4}{*}{\includegraphics[width=16mm, height=12mm]{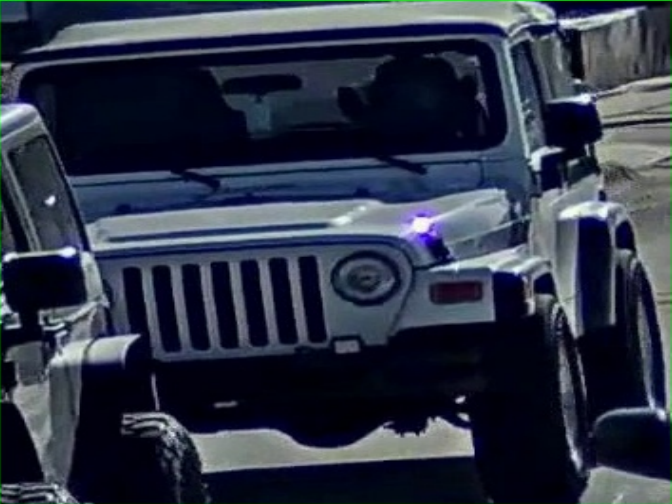}}   &Normal\\
        &   &   &\textbf{2}  &\textbf{2,063}   &\textbf{84.31}  &    &Normal\\
        &   &   &5  &628   &70.10  &    &Normal\\
        &   &   &10  &386   &40.20  &    &Abnormal\\
    \midrule
    \multirow{4}{*}{S05/c034} &\multirow{4}{*}{342} &\multirow{4}{*}{10} &1  &2,908   &95.27  &\multirow{4}{*}{\includegraphics[width=16mm, height=12mm]{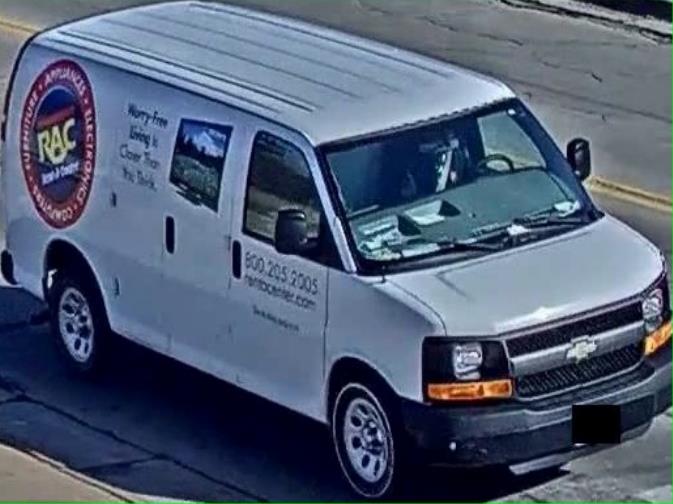}}   &Normal\\
        &   &   &\textbf{2}  &\textbf{1,706}   &\textbf{84.52}  &    &Normal\\
        &   &   &5  &711   &40.24  &    &Abnormal\\
        &   &   &10  &303   &11.24  &    &Abnormal\\
    \midrule
    \multirow{4}{*}{S05/c035} &\multirow{4}{*}{347} &\multirow{4}{*}{10} &1  &1,693   &96.84 &\multirow{4}{*}{\includegraphics[width=16mm, height=12mm]{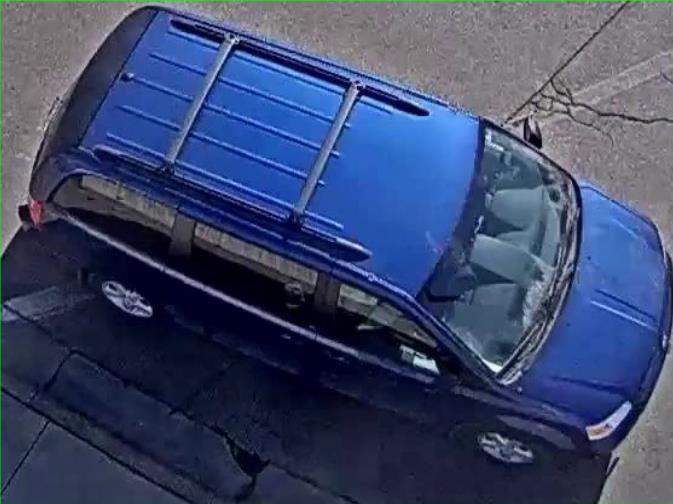}}   &Normal\\
        &   &   &\textbf{2}  &\textbf{1,001}   &\textbf{85.26}  &    &Normal\\
        &   &   &5  &351   &50.56  &    &Normal\\
        &   &   &10  &185   &11.05  &    &Abnormal\\
    \midrule
    \multirow{4}{*}{S05/c036} &\multirow{4}{*}{343} &\multirow{4}{*}{10} &1  &2,768   &87.83 &\multirow{4}{*}{\includegraphics[width=16mm, height=12mm]{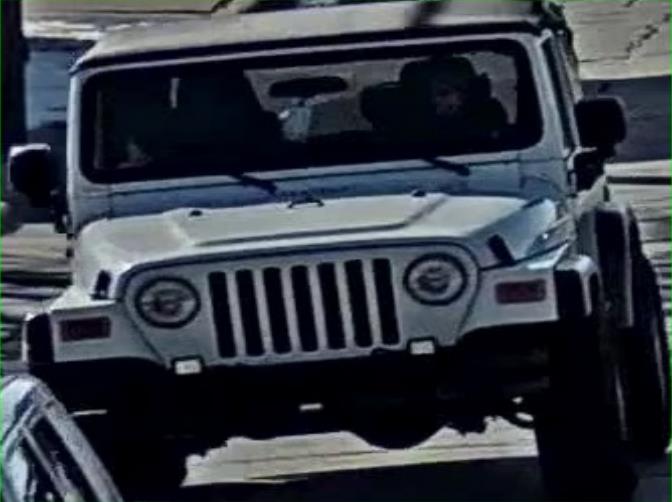}}   &Normal\\
        &   &   &\textbf{2}  &\textbf{1,805}   &\textbf{77.39}  &    &Normal\\
        &   &   &5  &1,024   &62.61  &    &Abnormal\\
        &   &   &10  &400   &39.13  &    &Normal\\
    \bottomrule
    \end{tabular}}
    \label{tab4}
\end{table}

\subsubsection{\textbf{Verification of Dynamic Annotation Data Loading}}
\label{subsub:store}
\begin{figure}
    \centering
    \includegraphics[scale=0.25]{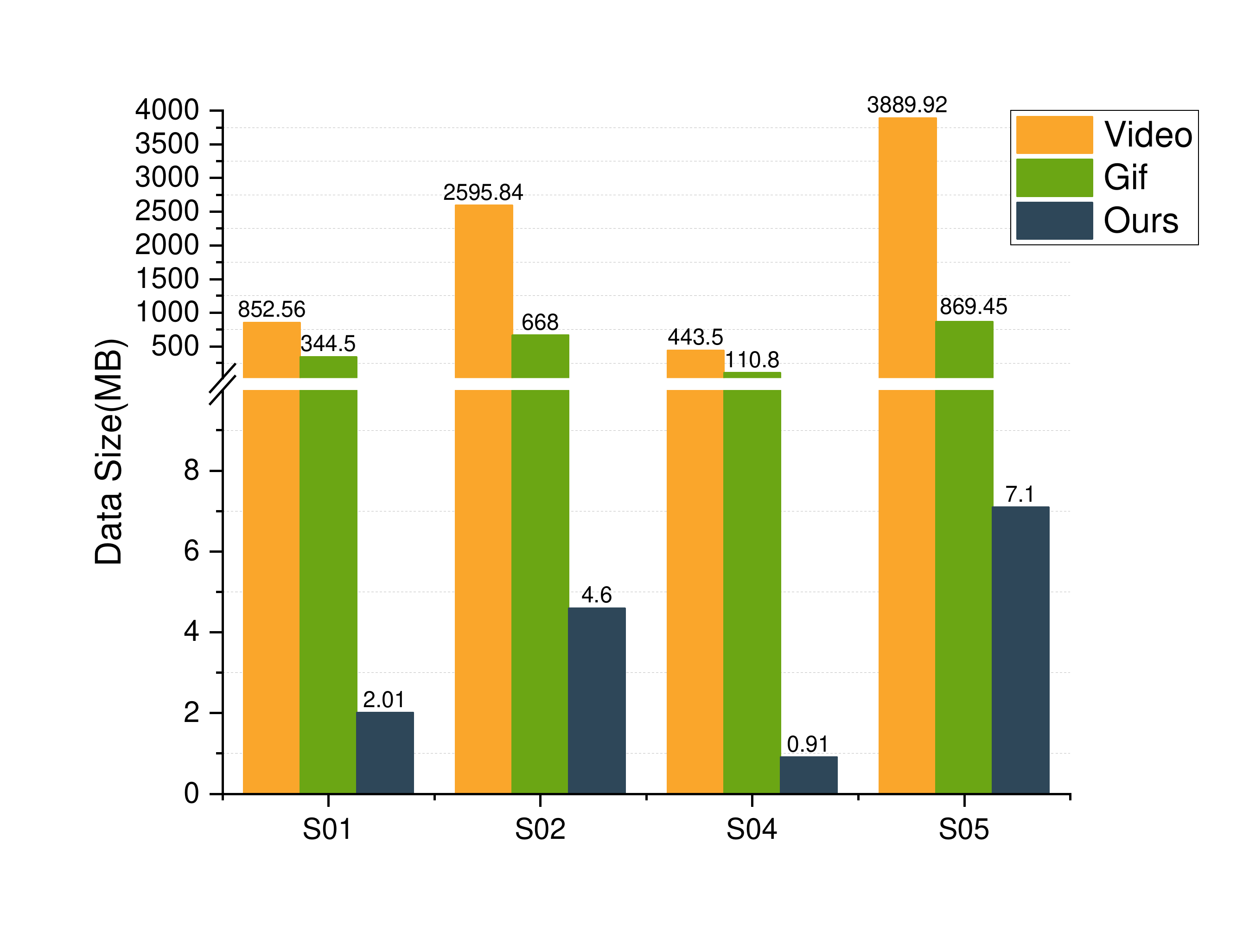}
    \caption{{Illustration of the storage resource requirement among different strategies.}}
    \label{tubiao_2}
\end{figure}

As mentioned in Sec. III, the proposed system employs a novel visualization strategy for annotated trajectory, which greatly reduces the storage requirements.   
%
This part compares the storage requirements between the proposed strategy and conventional methods that visualize the trajectories via GIF or video files.
The results are shown in table~\ref{tab5} and Fig.~\ref{tubiao_2}. 
%
As is shown, the proposed strategy reduces around $99\%$ of storage requirement compared with the conventional method. 
In addition, the proposed strategy can provide much better perceptual qualities than the method employing GIF format.

\begin{table}
    \centering
    \caption{Evaluation of Dynamic Annotation Data Loading Strategy}
    \scalebox{0.7}{
    \begin{tabular}{ccccccc}
    \toprule
        &Name     &Length (s)   &Original video size (MB) &\multicolumn{2}{c}{Trajectory visual size (MB)}  &\textbf{Ours (MB)}  \\
        &   &   &   &Video  &Gif    &\\
    \midrule
    \multirow{2}{*}{S01} &c003    &199 &151 &1,157  &376 &\textbf{2.94}\\
    &c005    &211 &55 &548  &313   &\textbf{1.08}\\
    \midrule
    \multirow{2}{*}{S02} &c006    &211 &292 &2,468  &587 &\textbf{5.09}\\
    &c009    &211 &205 &2,724 &749   &\textbf{4.10}\\
    \midrule
    \multirow{4}{*}{S04} &c030    &63 &45 &202  &68  &\textbf{0.96}\\
    &c033    &35 &43 &368 &97    &\textbf{0.76}\\
    &c034    &41 &53 &605 &156 &\textbf{0.83}\\
    &c036    &36 &55 &599 &122 &\textbf{1.08}\\
    \midrule
    \multirow{16}{*}{S05} &c010    &407 &473 &1,761  &401    &\textbf{2.45}\\
    &c016    &394 &425 &1,843 &533    &\textbf{3.93}\\
    &c017    &387 &552 &1,823 &374    &\textbf{6.37}\\
    &c018    &384 &593 &2,836 &728    &\textbf{15.20}\\
    &c021    &400 &435 &1,331 &387    &\textbf{4.55}\\
    &c022    &427 &611 &2,560 &544    &\textbf{6.00}\\
    &c023    &425 &1,004 &3,533 &893    &\textbf{9.21}\\
    &c025    &428 &546 &1,731 &668    &\textbf{10.40}\\
    &c026    &417 &675 &4,567 &983    &\textbf{3.69}\\
    &c027    &384 &600 &4,506 &936    &\textbf{3.46}\\
    &c028    &382 &434 &4,291 &1,055    &\textbf{8.89}\\
    &c029    &354 &768 &6,871 &1,085    &\textbf{6.75}\\
    &c033    &340 &553 &8,602 &1,751    &\textbf{9.22}\\
    &c034    &342 &679 &7,987 &1,444    &\textbf{6.28}\\
    &c035    &347 &346 &3,164 &931    &\textbf{8.77}\\
    &c036    &343 &429 &4,833 &1,198    &\textbf{8.50}\\
      \midrule
     &Average  &- &- &  2,954.58 &682.46   &\textbf{5.44}\\
    \bottomrule
    \end{tabular}
    }
    \label{tab5}
\end{table}

\section{Conclusion}
In this paper, we propose an efficient, accurate, and low-cost semi-automatic MTMCT data annotation system. 
The system first employs state-of-the-art learning-based object detection, re-ID and MOT algorithms to automatically generate single-camera trajectory features.
In the following manual cross-camera trajectory matching process, the system makes full use of side information including timestamp, camera location and background scene to provide users a reliable recommendation list.
This would significantly improve the matching accuracy and efficiency.
In addition, plenty of efforts have been devoted to system optimization.
To be specific, the dynamic annotation data loading saves quite a storage requirement comparing with the conventional annotation data visualization method.
And the implemented flexible deployment strategy allows multi-person online collaboration.
The proposed system considerably contributes to MTMCT studies, allowing researchers to establish large-scale datasets with acceptable human labor and time cost.
%
%

\section*{Acknowledgment}
This work was supported in part by the National Natural Science Foundation of China (Grant 61871270), in part by the Shenzhen Natural Science Foundation under Grants JCYJ20200109110410133 and 20200812110350001.

\ifCLASSOPTIONcaptionsoff
  \newpage
\fi

\bibliographystyle{ieeetr}
\bibliography{ref}

%

\begin{IEEEbiography}{Michael Shell}
Biography text here.
\end{IEEEbiography}

\begin{IEEEbiographynophoto}{John Doe}
Biography text here.
\end{IEEEbiographynophoto}


\begin{IEEEbiographynophoto}{Jane Doe}
Biography text here.
\end{IEEEbiographynophoto}




\end{document}